\documentclass{article}
\usepackage{spconf,amsmath,graphicx}

\usepackage{algorithm}
\usepackage{algorithmic}
\usepackage{array}
\usepackage{color}
\usepackage{diagbox}
\usepackage{gensymb}
\usepackage{threeparttable}

\usepackage{amsfonts}
\usepackage{bm}

\usepackage{amssymb}
\usepackage{multirow}

\newcommand{\ie}{\emph{i.e. }}

\newcommand{\etal}{\emph{et al. }}

\newcommand{\vct}[1]{\boldsymbol{#1}} 
\newcommand{\mat}[1]{\boldsymbol{#1}} 

\usepackage{setspace}

\title{SELFGAIT: A SPATIOTEMPORAL REPRESENTATION LEARNING METHOD FOR SELF-SUPERVISED GAIT RECOGNITION}

\name{Yiqun Liu$^1$ \quad \quad Yi Zeng$^1$ \quad \quad Jian Pu$^2$ \quad \quad Hongming Shan$^2$ \quad \quad Peiyang He$^1$ \quad \quad Junping Zhang$^1$$\dag$ \thanks{$\dag$ Corresponding author}}
\address{$^1$Shanghai Key Lab of Intelligent Information Processing, School of Computer Science\\
	$^2$Institute of Science and Technology for Brain-inspired Intelligence \\
	 Fudan University, China, 200433\\
	 \{yqliu17, yzeng20, jianpu, hmshan, pyhe19, jpzhang\}@fudan.edu.cn}
\begin{document}
%
\maketitle

%
%
\begin{abstract}
Gait recognition plays a vital role in human identification since gait is a unique biometric feature that can be perceived at a distance. Although existing gait recognition methods can learn gait features from gait sequences in different ways, the performance of gait recognition suffers from insufficient labeled data, especially in some practical scenarios associated with short gait sequences or various clothing styles. It is unpractical to label 
the numerous gait data. In this work, we propose a self-supervised gait recognition method, termed SelfGait, which takes advantage of the massive, diverse, unlabeled gait data as a pre-training process to improve the representation abilities of spatiotemporal backbones. 
Specifically, we employ the horizontal pyramid mapping (HPM) and micro-motion template builder (MTB) as our spatiotemporal backbones to capture the multi-scale spatiotemporal representations. 
Experiments on CASIA-B and OU-MVLP benchmark gait datasets demonstrate the effectiveness of the proposed SelfGait compared with four state-of-the-art gait recognition methods. The source code has been released at \textcolor{magenta}{\emph{https://github.com/EchoItLiu/SelfGait}}. 
\end{abstract}

\section{Introduction} \label{sec:intro}
Unlike other commonly-used biometrics such as face and fingerprint, gait is the unique biometric feature that can be used for identifying humans at a far distance. Gait recognition has broad applications in crime prevention and social security. In real-world scenarios, however, gait recognition suffers from exterior factors such as coat-wearing and carrying a bag. The camera viewpoints also result in dramatic changes in gait appearance, which makes gait recognition considerably challenging.

Currently, various deep learning-based methods have been proposed to tackle these issues. They can be roughly grouped into two categories. One approach is to compress all gait silhouettes into one image or a template for gait information \cite{he2018multi, takemura2017input, wu2016comprehensive, shiraga2016geinet}. Wu \etal firstly introduced the CNN-based method for capturing the deep features of gait patterns from gait energy images (GEI) \cite{wu2016comprehensive}. Besides, He \etal proposed multi-task generative adversarial networks (MGANs) and multi-channel gait template named period energy image (PEI) to learn more view-specific representations and extract the temporal information simultaneously \cite{he2018multi}. The second category of gait recognition directly encodes the spatial and temporal representations from the original gait silhouette sequences \cite{wolf2016multi, chao2019gaitset, fan2020gaitpart, song2019gaitnet, liao2017pose}. For example, GaitNet \cite{song2019gaitnet} utilized auto-encoder to extract the features from raw silhouette sequences, followed by employing a long short-term memory (LSTM) to model the temporal variations of the gait sequence. To enhance the flexibility of gait recognition, GaitSet regarded gait as a set instead of a sequence so that the number of gait samples could be enriched~\cite{chao2019gaitset} 
and obtained a SOTA performance. More recently, GaitPart \cite{fan2020gaitpart} captured the discriminative spatiotemporal characteristics from each part of the body using the micro-motion template builder (MTB). 

It is worth noting that many silhouette-based methods assume the whole human body as a unit to establish the spatiotemporal representations \cite{fu2019horizontal}. Further, gait recognition methods suffer from the insufficiency of the training data because the acquisition of the labeled gait images is difficult and expensive in practical scenarios because of lots of occlusion~\cite{Wu_2020_CVPR}and dressing~\cite{Yu_2020_CVPR}. Therefore, the numerous unlabeled gait images cannot be utilized for training the gait model even if they potentially contain rich information. To make use of these unlabeled sets to improve recognition performance, we propose the SelfGait algorithm by adopting the self-supervised framework named ``bootstrap your own latent'' (BYOL)~\cite{grill2020bootstrap} for pre-training with the unlabeled samples to boost the representation ability of spatiotemporal backbones. More specifically, 
we use the BYOL with deliberately designed online networks and target networks. These two networks based on the spatiotemporal pretext task are to separately predict the two different gait identity features of one person, and can be optimized by {\it cosine} similarity loss to promote the representation abilities of spatiotemporal backbones. The horizontal pyramid mapping (HPM) of GaitSet~\cite{chao2019gaitset} and micro-motion template builder (MTB) ~\cite{fan2020gaitpart} of GaitPart~\cite{fan2020gaitpart} are selected as spatiotemporal backbones of our model. Quantitative and ablative experiments demonstrate that the proposed SelfGait can achieve better performance by conducting the pre-training step by unlabeled gait images and then fine-tuning with a portion of the labeled training set, compared with other four SOTA gait recognition methods.

\section{Method} \label{sec:method}
\begin{figure}[t]
\centering
\includegraphics[width=1.0\linewidth, clip=true]{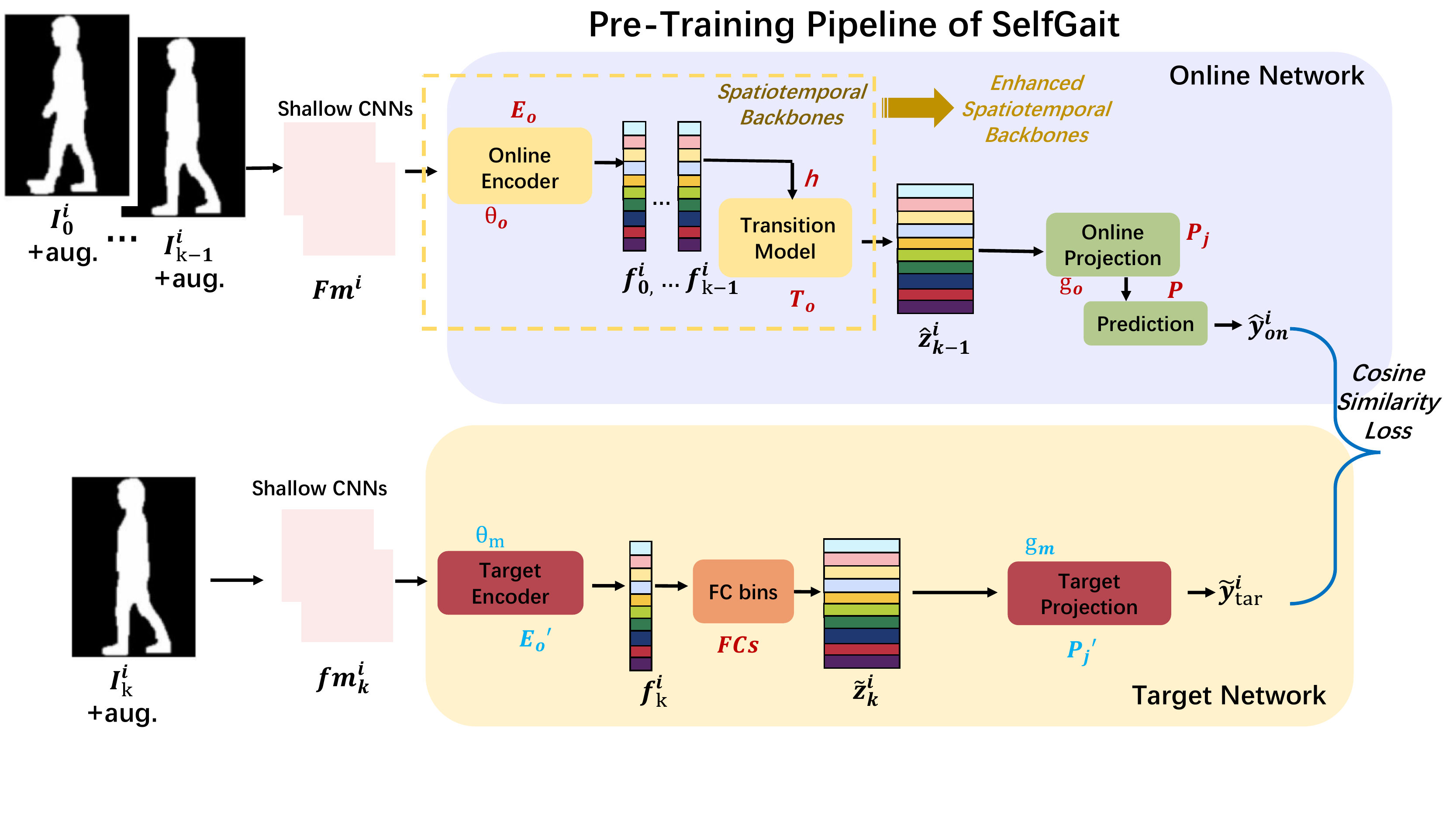}
\caption{{\bf The pre-training pipeline of SelfGait} for strengthening the representation abilities of spatiotemporal backbones that are Online Encoder and Transition model according to the golden arrow pointer. For one sample in this pipeline, the $0\sim k-1$ frames of gait sequence from one person and its $k$-th frame separately pass the Online network and Target network to output the several identity features $\hat{\vct{y}}^i_{on}$ and $\tilde{\vct{y}}^{i}_{tar}$.}
\label{pipeline}
\end{figure}
\subsection{Preliminary}
As shown in Fig.~\ref{pipeline}, the pre-training pipeline of our SelfGait consists of an Online network and a Target network~\cite{grill2020bootstrap}. The pre-training process of SelfGait is to strengthen the feature extraction ability of \emph{Spatiotemporal Backbones} based on the specified pretext task, which is set for predicting two identity features of one
person through the two network branches respectively and then calculate the cosine similarity between features. Meanwhile, given a dataset of $N$ people with identities $y_i, i\in{1,2, ..., N}$, all silhouettes in one gait sequence of a given person is defined as $\mathcal{\mat{I}}^i=\{\mat{I}^i_{t}|t=0,1,2,...,k\}$, where $k=30$ is the number of frames. We feed the $0\sim k-1$ frames of $\mathcal{I}^i$ and the $k$-th frame of $\mathcal{\mat{I}}^i$ into different shallow CNNs and acquire the feature maps $\mat{Fm}^i$ and $\mat{fm}^i_{k}$.
\subsection{Online and Target Networks}
\begin{figure}[t]
\centering
\includegraphics[width=0.80\linewidth, clip=true]{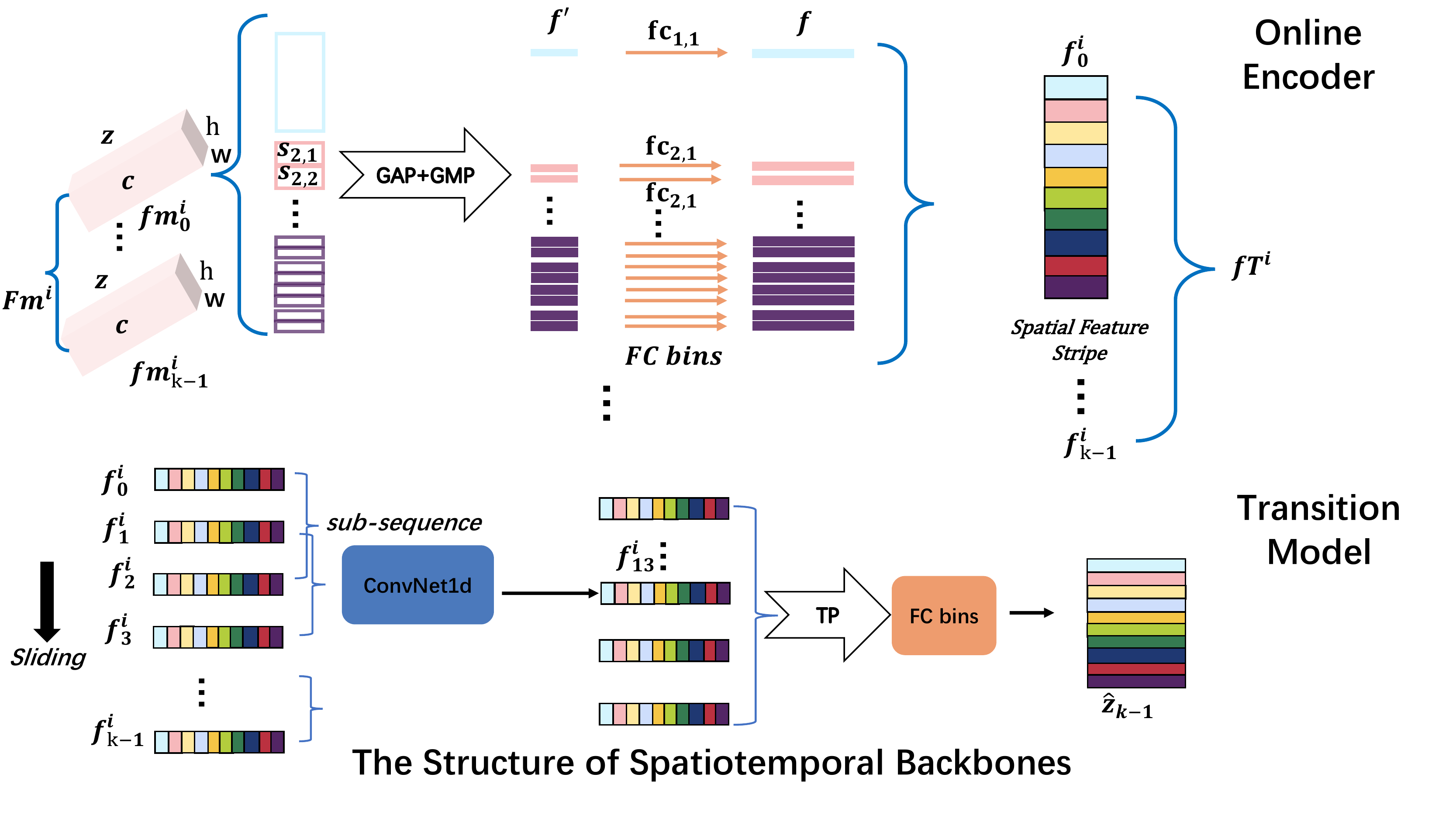}
\caption{From top to bottom: {\bf the detail structures of spatiotemporal backbones of SelfGait} for capturing the spatiotemporal representations. The FC bins consist of many independent full connected layers for acquiring the multi-scale feature stripe $\mat{fT}^{i}$ for $i$-th person.}
\label{backbones}
\end{figure}

{\bf Online Networks}\quad The feature maps $\mat{Fm}^i$ forward the upper branch for conducting the task of predicting the identity features $\hat{\vct{y}}^{i}_{on}$. The pipeline of this task is formulated by $\hat{\vct{y}}^{i}_{on} = P(P_j(T_o(E_o(\mat{Fm}^i))))$. Specifically, $\mat{Fm}^i$ passes the Online Encoder $E_o$ that is a horizontal pyramid mapping (HPM)~\cite{chao2019gaitset} for capturing the multi-scale spatial features. On the top of Fig~\ref{backbones}, more concretely, every feature frame of $\mat{Fm}^i$ orderly enters the HPM and is divided into $2^{s-1}$ patches on the height dimension, \ie $n = \sum_{s=1}^S2^{s-1}$ patches in total, where $S=5$ is the number of scales. Subsequently, the global and average pooling are performed around all patches, each of which is fed into independent FCs severally for outputting the multi-scale spatial feature stripe $\mat{fT}^{i} = \{\vct{f}^{i}_{0},...,\vct{f}^{i}_{k-1}\}$. As shown on the bottom of Fig~\ref{backbones}, the $\mat{fT}^{i}$ is then passed the transition model $T_{o}$ for learning the temporal features. That is to say, we slice the multi-scale time sub-sequences from the neighbor fore-and-aft $r = 1$ frames of every element of $\mat{fT}^{i}$ and employ a micro-motion template builder (MTB)~\cite{fan2020gaitpart} for performing the temporal convolutional network (TCN) on all sub-sequences. Then, a pooling operation on temporal dimension (TP) is performed on convoluted temporal features. Finally, the spatiotemporal representations $\hat{\vct{z}}_{k-1}\in\mathbb{R}^{n*d_1}$ are acquired by mapping operation from FC bins, where $n$ is the number of scale and $d_1$ is dimension of out channel. The last two layers of upper branch are Online Projection $P_{j}$ (Conv+Relu+BN) and Prediction $P$ are devoted to output the identity features of Online Network $\hat{\vct{y}}^i_{on}\in\mathbb{R}^{n*d_2}$, where $d_2$ is feature dimension of $\hat{\vct{y}}^i_{on}$. It is worth mentioning that batch normalization (BN) operation from the projecting head of $P_{j}$ can reassign the features to prevent from mode collapse.

{\bf Target Networks}\quad The pipeline of target branch is formulated by $P_{j}'(FCs(E_{o}'(\mat{fm}^i_{k})))$, where the Target Encoder $E_{o}'$ and Target Projection $P_{j}'$ have same structure of Online Encoder and Online Projection respectively. The function of $FCs$ shared with the parameters of counterpart from Transition Model maps the spatial feature $\tilde{\vct{z}}^{i}$ output from $E_{o}'$ into same dimension of $\hat{\vct{z}}^{i}_{k}$. Finally, the identity features of Target Network $\tilde{\vct{y}}^{i}_{tar}$ is acquired by $P_{j}'$.

\begin{figure}[htbp]
\centering
\includegraphics[width=0.80\linewidth, clip=true]{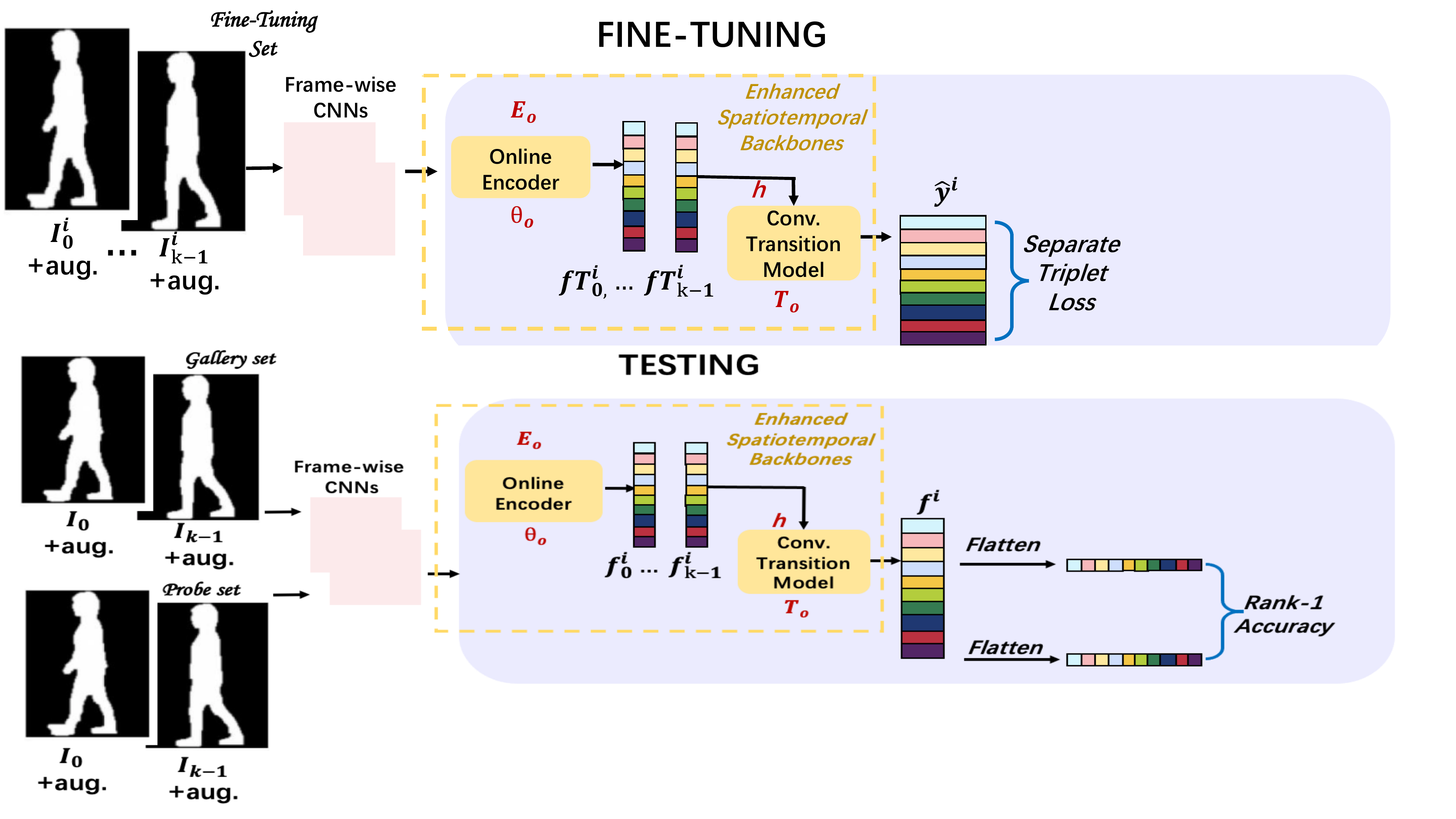}
\caption{{\bf The fine-tuning and testing pipelines of SelfGait} consist of \emph{Enhanced Spatiotemporal Backbones} that are obtained by pre-training.}
\label{FineTune}
\end{figure}

\subsection{Pre-training, Fine-tuning and Testing}
We perform the pre-training process upon the self-supervised (SS) SelfGait framework illustrated in Fig.~\ref{pipeline} to enhance the representative ability of \emph{Spatiotemporal Backbones} and then conduct the fine-tuning/testing in virtue of the \emph{Enhanced Spatiotemporal Backbones}. The details are discussed below:

\noindent {\bf Pre-training}\quad The SS SelfGait framework is optimized by the cosine similarity loss formulated by:
\begin{equation}
\mat{L}_{cos} = \left(\frac{\tilde{\vct{y}}^{i}_{tar}}{||\tilde{\vct{y}}^{i}_{tar}||_2}\right)^{T}\left(\frac{\hat{\vct{y}}^i_{on}}{||\hat{\vct{y}}^i_{on}||_2}\right)
\end{equation} where $\tilde{\vct{y}}^{i}_{tar}$ and $\hat{\vct{y}}^i_{on}$ are the identity features predicted from Target Network and Online Network severally.  

\noindent {\bf Fine-Tuning and Testing}\quad In Fig~\ref{FineTune}, we fine-tune the spatiotemporal components and optimize it by Batch All~($BA_+$) triplet loss~\cite{2017arXiv17Tri} defined below, where $\xi$ is the margin between intraclass/interclass distances $D_{\alpha, \beta}$ and $D_{\alpha, \gamma}$.  
\begin{align}
\label{eq:loss}
L(r) = ReLU(\xi+D_{\alpha, \beta}-D_{\alpha, \gamma}),
\end{align}


\section{experiments}
\begin{table*}[]
\centering
\resizebox{250pt}{!}{
\scriptsize
\begin{tabular}{l|l|cccccccccccc}
\hline
\multicolumn{2}{l|}{Gallery NM\#1-4} & \multicolumn{11}{c|}{0 - 180}                                                                  & \multicolumn{1}{l}{\multirow{2}{*}{Mean}} \\ \cline{1-13}
\multicolumn{2}{l|}{Probe}           & 0    & 18   & 36   & 54   & 72   & 90   & 108  & 126  & 144  & 162  & \multicolumn{1}{c|}{180} & \multicolumn{1}{l}{}                      \\ \hline
\multirow{6}{*}{NM \#5-6} & GaitSet \cite{chao2019gaitset}  & 90.8 & 97.9 & 99.4 & 96.9 & 93.6 & 91.7 & 95.0 & 97.8 & 98.9 & 96.8 & 85.8                     & 95.0                                      \\
                          & GaitNet \cite{song2019gaitnet}  & 91.2 & 92.0 & 90.5 & 95.6 & 86.9 & 92.6 & 93.5 & 96.0 & 90.9 & 88.8 & 89.0                     & 91.6                                      \\
                          & GaitPart \cite{fan2020gaitpart} & 94.1 & 98.6 & 99.3 & 98.5 & 94.0 & 92.3 & 95.9 & 98.4 & 99.2 & 97.8 & 90.4                     & 96.2                                      \\
                          & SG-2     & 84.8  & 88.4  & 93.8  & 96.1  & 88.8  & 85.0  & 86.2  & 89.4  & 93.0  & 90.8 & 84.0                      & 89.1                                       \\
                          & SG-4     & 88.4  & 93.2  & 95.2  & 96.0  & 91.4  & 90.8  & 92.2  & 93.6  & 95.2  & 92.2  & 90.2                     & 92.6                                       \\
                          & SG-6     & 90.4  & 93.8  & 96.8  & 96.7  & 92.0  & 92.2  & 92.2  & 94.6  & 96.8  & 94.8  & 88.6                      & 93.5                                      \\ \hline

\multirow{6}{*}{BG \#1-2} & GaitSet \cite{chao2019gaitset} & 83.8 & 91.2 &  91.8 & 88.8 & 83.3 & 81.0 & 84.1 & 90.0 & 92.2 & 94.4 & 79.0                     & 87.2                                      \\
                          & GaitNet \cite{song2019gaitnet} & 83.0 & 87.8 & 88.3 & 93.3 & 82.6 & 74.8 & 89.5 & 91.0 & 86.1 & 81.2 & 85.6                     & 85.7                                      \\
                          & GaitPart \cite{fan2020gaitpart} & 89.1 & 94.8 & 96.7 & 95.1 & 88.3 & 94.9 & 89.0 & 93.5 & 96.1 & 93.8 & 85.8                     & 91.5                                      \\
                          & SG-2     & 76.9  & 83.1  & 83.1  & 86.9  & 76.7  & 71.5  & 75.0  & 83.3  & 86.7  & 86.3  & 80.8                      & 81.3                                       \\
                          & SG-4     & 85.6  & 89.2  & 91.9  & 89.0  & 82.9  & 81.3  & 83.5  & 87.9  & 86.5  & 90.2  & 82.9                      & 86.4                                       \\
                          & SG-6    & 90.6  & 91.9  & 94.1  & 91.2  & 87.9  & 84.5  & 86.4  & 90.6  & 90.6  & 93.3  & 90.0                      & {\bf 90.1}                                       \\ \hline
\multirow{6}{*}{CL \#1-2} & GaitSet \cite{chao2019gaitset} & 61.4 & 75.4 & 80.7 & 77.3 & 72.1 & 70.1 & 71.5 & 73.5 & 73.5 & 68.4 & 50.0                     & 70.4                                      \\
                          & GaitNet \cite{song2019gaitnet} & 42.1 & 58.2 & 65.1 & 70.7 & 68.0 & 70.6 & 65.3 & 69.4 & 51.5 & 50.1 & 36.6                     & 58.9                                      \\
                          & GaitPart \cite{fan2020gaitpart} & 70.7 & 85.5 & 86.9 & 83.3 & 77.1 & 72.5 & 76.9 & 82.2 & 83.8 & 80.2 & 66.5                     & 78.7                                      \\
                          & SG-2     & 69.4  & 78.8 & 76.9 & 75.4 & 69.2 & 65.0 & 71.3 & 72.5 & 75.2 & 72.7 & 68.1                      & 72.2                                       \\
                          & SG-4     & 71.2  & 84.6  & 82.3  & 80.8  & 76.3  & 74.4  & 80.0  & 85.2  & 79.0  & 79.6  & 75.8                      & 79.0                                       \\
                          & SG-6    & 76.7  & 82.7  & 88.3  & 84.2  & 80.4  & 78.5  & 84.6  & 83.9  & 83.5  & 80.6  & 71.5                      & {\bf 81.3}                                       \\ \hline
\end{tabular}
}
\caption{Averaged rank-1 accuracies on CASIA-B, excluding identical-view cases.
}
\label{casia}
\end{table*}

\begin{table}[]
\centering
\resizebox{150pt}{!}{
\scriptsize
\setlength{\tabcolsep}{1mm}{
\begin{tabular}{c|c|c|c|c}
\hline
\multirow{1}{*}{\centering Probe} & \multicolumn{4}{c}{Gallery All 14 views}             \\ \cline{2-5} 
                       & GEINet \cite{shiraga2016geinet} & GaitSet \cite{chao2019gaitset} & GaitPart \cite{fan2020gaitpart} & SG-2 \\ \hline
0                      & 11.4   & 79.5    & 82.6             & 85.1        \\ \hline
15                     & 29.1   & 87.9    & 88.9            & 89.3        \\ \hline
30                     & 41.5   & 89.9    & 90.8            & 92.0        \\ \hline
45                     & 45.5   & 90.2    & 91.0            & {\bf 94.3}        \\ \hline
60                     & 39.5   & 88.1    & 89.7             & 89.1        \\ \hline
75                     & 41.8   & 88.7    & 89.9            & 90.2        \\ \hline
90                     & 38.9   & 87.8    & 89.5           & 90.9        \\ \hline
180                    & 14.9   & 81.7    & 85.2            & 87.4        \\ \hline
195                    & 33.1   & 86.7    & 88.1            & {\bf 91.8}        \\ \hline
210                    & 43.2   & 89.0    & 90.0            & 89.3        \\ \hline
225                    & 45.6   & 89.3    & 90.1            & 88.7        \\ \hline
240                    & 39.4   & 87.2    & 89.0             & 90.8        \\ \hline
255                    & 40.5   & 87.8    & 89.1            & 91.6        \\ \hline
270                    & 36.3   & 86.2    & 88.2             & 87.7        \\ \hline
mean                   & 35.8   & 87.1    & 88.7           & {\bf 89.9}        \\ \hline
\end{tabular}
}
}
\caption{Averaged rank-1 accuracies on OU-MVLP, excluding identical-view cases.}
\label{oumvlp}
\end{table}

\subsection{Implementation Details}
We conducted experiments on two benchmarked gait datasets with the full batch, including the 16 gait sequences of 8 persons. The \textbf{CASIA-B} dataset~\cite{yu2006framework} contains 124 subjects 3 walking conditions and 11 views. We use the first 74 subjects for training and the rest 50 subjects for testing. In the test sets, the first four sequences of the NM condition~(i.e., NM \#1-4) are kept in the gallery, and the remaining six sequences were divided into three probe subsets, i.e., the NM subset condition \#5-6, the BG subset containing BG \#1-2 and the CL subset containing CL \#1-2. and 2) \textbf{OU-MVLP} dataset~\cite{takemura2018multi} has 10,307 subjects, 14 views per subject. We strictly followed the setting used in~\cite{chao2019gaitset} for a fair comparison. In detail, the input is a set of aligned silhouettes in size of $64\times 44$, and we adopt Adam as our optimizer with a fixed learning rate of $10^{-4}$. All models are trained with 4 GPUs with 1) 80K iterations for CASIA-B and 2) and 150K iterations with learning rate decay for OU-MVLP for strengthening the representation abilities of spatiotemporal backbones.

\subsection{Main Results}
{\bf CASIA-B} \quad As shown in Table~\ref{casia}, we prove the effectiveness of promoting spatiotemporal backbones of SelfGait via the self-supervised representation learning using the training set discussed in 3.1. We select a proportion of samples from OU-MVLP as a pre-training set only and regard the $40\%$ samples of the training set as a fine-tuning set. The sample ratios between pre-training set and training set for SelfGait are set roughly $2$:$1$ (SG-2), $4$:$1$ (SG-4), and $6$:$1$ (SG-6) separately. The testing set is the same for all models, and all the cross-view and cross-walking-condition cases are included in the comparison scope. We first observe that SG-6 obtains preferable performance on CL \#1-2, i.e., the averaged accuracy of $81.3\%$, and the accuracy of SG-6 on BG \#1-2  is higher than GaitSet in the same case. It indicates that the performance of SelfGait can be enhanced by pre-training with the extra unlabeled samples. 2) It is worth mentioning that a higher performance enhancement from SG-2 to SG-6 is obtained. One possible explanation is that the SG-2 is fine-tuned by a few labeled training sets and conducts the pre-training with insufficient unlabeled samples, leading to inferior performance. Meanwhile, the promotion of SG-6 possibly benefits from the strong representative capacity of \emph{Enhanced Spatiotemporal Backbones} obtained by pre-training with abundant unlabeled samples. However, the performance of SelfGait in NM \#5-6 is worse than the baseline models and BG/CL \#1-2 in same cases. One possible reason is that the features from NM subjects of OU-MVLP in the pre-training process do not possess high discrimination, and therefore achieve an inferior contribution to boosting the representation, compared with the other two groups (BG \#1-2 and CL \#1-2) in the same case. Finally, both GaitSet and GaitPart have better performances than GaitNet because they can capture the high-level representations via the multi-scale pyramid backbones. Specifically, GaitPart utilized a micro-motion module to capture the multi-scale and short-range discriminative spatiotemporal (ST) features, while GaitNet only uses the auto-encoder composed of the LSTMs. The analysis above verifies the effectiveness of that representation learning by carrying out the pre-training on spatiotemporal backbones for gait recognition.

\noindent {\bf OU-MVLP} \quad  To verify the generalization of our SelfGait, we evaluate SelfGait on the worldwide largest public gait dataset called OU-MVLP. As shown in Table~\ref{oumvlp}, SG-2  meets a new state-of-the-art under various cross-view conditions. Compared with the results of {\bf CASIA-B}, the mean accuracy of SelfGait in all views obtains preferable performance because the OU-MVLP, including the subjects, have more occlusion and comprehensive exterior factors, which makes the testing task more challenge.

\subsection{Ablation Study}
To verify the effectiveness of each component in SelfGait, we perform ablation studies with various settings on CASIA-B, as shown in Table~\ref{ablation}. The ablation studies include replacing the horizontal pyramid mapping (HPM) with plain CNNs (SG-6 $w/o$ (without) HPM), replacing the micro-motion template builder (MTB) with TCNs (SG-6 $w/o$ MTB), and ignoring the pre-training process and directly fine-tuning to verify the availability of self-supervised framework  (SG-6 $w/o$ PT). The experiment results and analysis are reported as follows.

\noindent {\bf Effectiveness of PT} \quad The first ablative study is to leave out the pre-training process, and directly train the spatiotemporal components shown in Fig.~\ref{FineTune} using $40\%$ training set. Obviously, accuracy decreases, which shows the effectiveness of strengthening the \emph{Enhanced Spatiotemporal Backbones} by self-supervised pre-training.

\noindent {\bf Effectiveness of HPM} \quad We then replace the HPM component with a three-layer CNNs. We noticed that SelfGait without HPM decreased performance significantly, which verifies the necessity of the spatial multi-scale pyramid for deep-learning-based gait recognition.

\noindent {\bf Effectiveness of MTB} \quad Finally, we replace the MTB with the plain TCNs. The accuracy of SelfGait without MTB loses degrades, which indicates that multi-scale temporal representation is equally essential for capturing the discriminative representations.

\begin{table}[t]
\centering
\scriptsize

    \begin{tabular}{|c|c|c|c|}
    \hline
    Components & NM & BG & CL \\
    \hline
    SG-6 $w/o$ PT & 85.5 & 79.8 & 73.6 \\   
    \hline
    SG-6 $w/o$ HPM & 73.3 & 63.8 & 52.7 \\
    \hline
    SG-6 $w/o$ MTB & 81.6 & 68.4 & 52.7 \\
    \hline
    SG-6 & 93.2 & 89.7 & 81.5 \\
    \hline
    \end{tabular}
  \caption{The ablation study conducted on \textbf{CASIA-B} using setting LT. The results are rank-1 accuracies averaged on all 11 views. }
  \label{ablation}
\end{table}

\section{Conclusions}
We have presented a novel self-supervised framework with spatiotemporal components to learn from the massive unlabeled gait images to boost gait recognition's performance. The SelfGait method is proposed to use the self-supervised framework as pre-training and then fine-tuned using the horizontal pyramid mapping (HPM) and the micro-motion template builder (MTB) as spatiotemporal components with only a few label samples. The experimental results on CASIA-B and OU-MVLP indicate that SelfGait achieves the preferable performance of gait recognition due to boosting the representation ability of spatiotemporal components, especially for the subjects with coat or jacket.

\section{Acknowledgement} \label{sec:acknow}

This work was supported in part by National Key Research and Development Program of China under Grant 2018YFB1305104, the Shanghai Municipal Science and Technology Major Project (No. 2018SHZDZX01), Shanghai Municipal Science and Technology Project (18DZ1200404) and ZJLab, and National Natural Science Foundation of China (NSFC 61673118).

\bibliographystyle{IEEEbib}
\bibliography{strings}

\end{document}